# A comparative analysis between Conformer-Transducer, Whisper, and wav2vec2 for improving the child speech recognition

Andrei Barcovschi
Ph.D. Student
University of Galway
Galway, Ireland
a.barcovschi1@universityofgalway.ie

Rishabh Jain
C3 Imaging Research Center
University of Galway
Galway, Ireland
rishabh.jain@universityofgalway.ie

Peter Corcoran
C3 Imaging Research Center
University of Galway
Galway, Ireland
peter.corcoran@universityofgalway.ie

*Abstract*— Automatic Speech Recognition (ASR) systems have progressed significantly in their performance on adult speech data; however, transcribing child speech remains challenging due to the acoustic differences in the characteristics of child and adult voices. This work aims to explore the potential of adapting state-of-the-art Conformer-transducer models to child speech to improve child speech recognition performance. Furthermore, the results are compared with those of self-supervised wav2vec2 models and semi-supervised multi-domain Whisper models that were previously finetuned on the same data. We demonstrate that finetuning Conformer-transducer models on child speech yields significant improvements in ASR performance on child speech, compared to the non-finetuned models. We also show Whisper and wav2vec2 adaptation on different child speech datasets. Our detailed comparative analysis shows that wav2vec2 provides the most consistent performance improvements among the three methods studied.

## I. INTRODUCTION

In the domain of Automatic Speech Recognition (ASR), several challenges persist, such as limited training data, untranscribed data, and difficulty in low-resource languages and children's speech. Recent research efforts have addressed some of these issues, leading to impressive ASR performance for adult speech, even achieving human-level performance [1]–[5]. However, progress in ASR for child speech has been slower, primarily due to the scarcity of annotated child-speech datasets required for effective training. Child speech datasets are challenging to collect and annotate, unlike adult speech data (as discussed in [6]). Moreover, inherent differences between adult and child voices, including pitch, linguistic and acoustic features, and pronunciation ability [7], [8], further hinder the performance of ASR models on child speech. The shorter vocal tract length and higher fundamental frequency [9] of children's voices also contribute to the complexity of accurately recognizing child speech.

The advantages and disadvantages of supervised and unsupervised ASR training approaches have been observed in recent developments, particularly in the context of child speech recognition. Unsupervised pretraining techniques like wav2vec2 [3] have shown significant improvements in child ASR [10]–[12]. However, their reliance on a finetuning stage with labeled data can limit their usefulness as they may overfit to specific datasets and not generalize well to diverse distributions. On the other hand, supervised learning approaches in child ASR [13]–[15] have explored transfer learning from adult to child speech [10], [13], [16], data augmentation methods [17]–[19], and weakly supervised training [15], [16], [20]. Recent findings [21], [22] indicate that supervised methods, involving pretraining on multiple datasets/domains, can enhance model robustness and generalization performance on unseen datasets. Nevertheless, each approach presents its trade-offs in terms of adaptability and scalability for diverse real-world speech recognition scenarios.

In this work, we use recent State-of-the-Art (SOTA) ASR models, Conformer-transducer for the task of child speech recognition. We also provide a comparative analysis of this model with our previously benchmarked results on wav2vec2 [23] and whisper [24]. Whisper is a supervised learning-based ASR system, which uses large amounts of labeled audio data. It uses weakly supervised pretraining beyond English-only speech recognition to be multilingual and multitask, showing great performance on different multilingual adult speech datasets [4]. The wav2vec2 is a self-supervised pretraining method for speech representations, leading to data-efficient finetuning for downstream ASR tasks. Conformer-transducer, combining CNNs and Transformers for end-to-end speech recognition, offers streaming capabilities and efficient long-range dependency modeling. While wav2vec2 is data-efficient and Whisper and Conformer-transducer excel in real-time processing, each model has unique strengths, making the choice dependent on factors like performance, model size, and application requirements. Since these models perform well on adult speech and gave SOTA results on widely used adult speech datasets, it was decided to use these models on different child speech datasets. We also finetune these models using different combinations of child speech datasets to see the subsequent speech recognition performance on different seen and unseen distributions of child speech datasets [25]–[27]. Our goal is to evaluate the efficacy of these methodologies in child speech analysis and determine their potential for enhancing child ASR technology and developing educational tools for children.

## II. MODEL DESCRIPTION

### A. Conformer-transducer [2]

The Conformer-transducer ASR model combines the benefits of both the transformer and CNN into a single architecture, namely the efficient global-level modeling of long-range dependencies in audio samples introduced by self-attention, and the finer-grained modeling of local dependencies enabled by convolutional kernels, respectively. The encoder network consists of a stack of Conformer blocks replacing the Transformer blocks [28]. A Conformer block consists of a feed-forward module followed by a multi-headed self-attention module, a convolution module, and finally another feed-forward module. Half-step residual connections always follow the feed-forward modules and a Layernorm is added as the last step in each block. The architecture of the Conformer encoder can be seen in Figure 1.

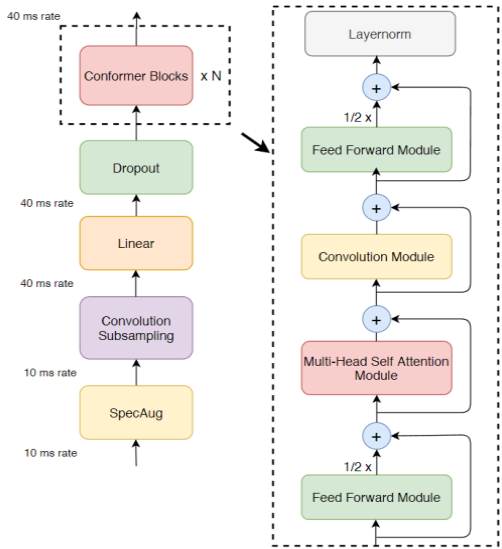

Figure 1: Conformer encoder model architecture [2].

Conformer-transducer models offer an improvement in WER for adult speech over the RNN-T and the Transformer-transducer architectures [2]. The Conformer-transducer uses the autoregressive transducer decoder, dropping the original simpler LSTM decoder. For the task of ASR, using the transducer decoder and transducer loss instead of the Connectionist Temporal Classification (CTC) [29] reduces incorrect spellings due to its autoregressive property, which implicitly models the inherent dependency between predicted output tokens, while CTC assumes that the output tokens are conditionally independent. However, this comes at the cost of larger GPU memory requirements for training and slower decoding speeds. Using a transducer approach introduces two new networks – the Decoder and the Joint model. The output of the Conformer's Encoder is inputted to the joint model, along with the autoregressive decoder model's output, and autoregressively produces a joint probability distribution over the known token vocabulary. At training time, the transducer loss is calculated over the output of the joint network.

### B. Whisper [4]

Whisper represents a significant advancement in weakly supervised pre-training, extending its capabilities to encompass multilingual and multitask scenarios beyond English-only speech recognition. Its strength lies in a vast dataset comprising 680,000 hours of labeled audio, where 117,000 hours cover 96 different languages, and an additional 125,000 hours include X→en translation data, where X is a non-English language and 'en' represents English translated data. Employing a transformer-based architecture with residual connections, the model handles an entire speech processing pipeline, encompassing transcription, translation, voice activity detection, alignment, and language identification. The Whisper model operates on 80-channel log-Mel spectrograms, with the encoder-decoder Transformer network featuring two convolutional layers, sinusoidal positional encoding, and a stacked set of Transformer blocks. The decoder uses learned positional embeddings and the same number of Transformer blocks as in the encoder. A comprehensive explanation of the Whisper architecture is available in [4].

### C. wav2vec2 [3]

wav2vec 2.0 is a speech recognition model based on self-supervised learning of speech representations through a two-stage architecture for pretraining and finetuning. The architecture comprises three key components: a CNN feature extractor, a transformer-based encoder, and a quantization module (see [3] for detailed information). During pretraining, the model is trained on a vast dataset of unlabeled speech data to acquire meaningful representations by capturing the temporal and spectral characteristics of speech. This is accomplished using a masked contrastive loss function. In the finetuning phase, the pretrained model is further trained on a smaller labeled dataset tailored for a specific downstream task. Here, the last layer of the pretrained model is substituted with a task-specific feed-forward layer, and the entire model is finetuned by minimizing the CTC loss [29] for ASR.

### D. Training Details

All models were trained on A6000 GPUs with 48GB of memory. The architectural parameters for Whisper, wav2vec2, and Conformer-transducer models utilized in this study are detailed in Table I.

TABLE I. ARCHITECTURE PARAMETERS FOR CONFORMER-TRANSDUCER[2], WHISPER[4], AND WAV2VEC2 [3] MODELS

| Models | Layers | Width | Heads | Learning Rate | Parameters |
|---|---|---|---|---|---|
| **Conformer-Transducer Models:** | | | | | |
| Small | 16 | 176 | 4 | 3.0 | 14M |
| Medium | 16 | 256 | 4 | 3.0 | 32M |
| Large | 17 | 512 | 8 | 3.0 | 120M |
| XLarge | 24 | 1024 | 8 | 3.0 | 600M |
| **Whisper Models:** | | | | | |
| Tiny | 4 | 384 | 6 | $1.5 \times 10^{-3}$ | 39M |
| Base | 6 | 512 | 8 | $1 \times 10^{-3}$ | 72M |
| Small | 12 | 768 | 12 | $5 \times 10^{-4}$ | 244M |
| Medium | 24 | 1024 | 16 | $2.5 \times 10^{-4}$ | 769M |
| Large | 32 | 1280 | 20 | $1.75 \times 10^{-4}$ | 1550M |
| **wav2vec2 Models:** | | | | | |
| Base | 12 | 768 | 8 | $5 \times 10^{-4}$ | 95M |
| Large | 24 | 1024 | 16 | $3 \times 10^{-4}$ | 317M |

### III. CORPUS DESCRIPTION

The Conformer-transducer pretrained models are trained on several thousand hours of English speech from diverse resources such as Librispeech, Fisher Corpus, Switchboard-1 Dataset, WSJ-0 and WSJ-1, National Speech Corpus, VCTK, VoxPopuli, Europarl, Multilingual Librispeech, Mozilla Common Voice, and People's Speech. The authors of Whisper [4] do not explicitly state the datasets used for training their models. Nonetheless, these trained models achieved SOTA results on various adult speech ASR datasets [4]. The wav2vec2-base model is pretrained with 960 hours of librispeech [30] and the wav2vec2-large model is pretrained with 60k hours of libri-light [31] datasets. In our study, we utilize three distinct child speech datasets and one adult speech dataset: MyST Corpus [25], PFSTAR dataset

[27], and CMU Kids dataset [26]. We maintain consistency with previous research wav2vec2 [23] and Whisper [24] to facilitate a direct comparison with the Conformer-transducer models.

*A. Dataset cleanup*

The cleaning process for the text labels involved removing abbreviations, punctuations, white spaces, and other non-alphanumeric characters, and converting all characters to lowercase. The audio data was modified to have a 16Khz sampling rate and 16-bit mono channel. For finetuning experiments, we used My Science Tutor (MyST) Corpus [25], an American English dataset. After cleaning and preparing this dataset according to [23], we divided 65 hours of clean child speech into two subsets: 55 hours for training and 10 hours for testing. Additionally, PFSTAR [27], a collection of words spoken by British English children, contributed 12 hours of audio, with 10 hours used for training and 2 hours used for testing. We also utilized CMU_Kids [26] corpus for validation-only, containing 9 hours of read-aloud sentences by children. While these datasets may not be extensive, they currently represent the best publicly available child speech datasets.

*B. Dataset Usage*

The datasets were divided according to their usage into a 'training' and an 'inference' set. This information is summarized in Table II.

TABLE II. DATASET USAGE

| Usage | Dataset | Duration |
|---|---|---|
| Finetuning (Training) | MyST_55h | 55 hours |
| | PFS_10h | 10 hours |
| Inference (Testing) | dev-clean | 9 hours |
| | MyST_test | 10 hours |
| | PFS_test | 2 hours |
| | CMU_test | 9 hours |

## IV. EXPERIMENTS AND RESULTS

*A. Codebase*

The Whisper implementation used is provided here[1]. The fairseq[2] implementation of wav2vec2 is used for finetuning experiments. The relevant information regarding model training, hyperparameters, graphs/metrics, checkpoints, and dataset availability are made available on our GitHub[3]. As for Conformer, we use its Nvidia's implementation for our experiments[4].

*B. Experiments*

The first set of Conformer-transducer experiments involved evaluating the original publicly available models on different child audio evaluation datasets mentioned in Table II without finetuning. The model sizes used were Small, Medium, Large, and XLarge as mentioned in Table I. For Whisper experiments, we use the Tiny, Base, Small, Medium, Large, and Large-V2 models. There are two versions of each model: one trained with multilingual data and one specifically for the English language only (indicated by '.en' in the name). The detailed list of experiments is mentioned in [23]. For wav2vec2 experiments, we use the 'Base' and 'Large' models which are pretrained with 960 hours of Librispeech data [30] and 60,000 hours of Libri-light data [31] respectively. Two models with the best performance from the first set of experiments are selected for further finetuning, namely, the models with the lowest WER. Finetuning included three experimental configurations of training data: MyST_55h, PFSTAR_10h, and MyST_55h+PFSTAR_10h combined.

The Conformer-transducer finetuning experiments on child speech involved finetuning only the feed-forward layers of all the encoder's Conformer blocks along with all layers of the decoder and joint networks of the base models. This decision was taken based on selecting the best result from preliminary experiments that tested different training hyperparameters and the finetuning of different combinations of layers of the Conformer-transducer large model, which can be found in the Appendix. The Adam optimizer was used with a base learning rate of 3.0 in combination with the Noam learning rate scheduler which linearly increased the learning rate for the first 40,000 steps before decaying exponentially. Greedy batch decoding was used as the token decoding strategy and for all experiments a unigram-based sentence-piece tokenizer with a vocabulary size of 1024 tokens was created for each unique finetuning dataset combination. The models were finetuned up to 500 epochs.

For whisper and wav2vec2 finetuning, the finetuning setup was kept consistent with previously reported results on Whisper [24] and wav2vec2 [23] approaches to provide a fair comparative analysis. We use a learning rate of $1 \times 10^{-5}$ for all Whisper finetuning experiments. The wav2vec2-base was finetuned with a learning rate of $1 \times 10^{-4}$, while wav2vec2-large was finetuned with a learning rate of $2.5 \times 10^{-5}$. Finetuning both approaches involves training the final layer of the models and freezing all others, as described by the respective authors. The Whisper model undergoes finetuning by minimizing the cross-entropy objective function, whereas wav2vec2 is finetuned by minimizing the CTC loss.

*C. Results and Discussions*

   *a)* **No-Funetuning Experiments**: Table III shows Word Error Rates (WERs) of original, non-finetuned Whisper, wav2vec2, and Conformer-transducer models on child speech evaluation datasets mentioned in Table II. No initial finetuning was performed over these models. A general trend of high WER on the MyST_test evaluation set can be observed across all the Whisper and Conformer-transducer models with most hovering around the 25% mark even for the much larger models. Only the wav2vec2 models perform better on MyST_test, displaying WERs that are approximately 10 points lower. We use these experiments as a baseline for further finetuning. The models with the lowest WER were chosen for providing executing further finetuning experiments with child speech.

---

[1] **Whisper Implementation:** https://github.com/huggingface/community-events/tree/main/whisper-fine-tuning-event

[2] **wav2vec2 Fairseq:** https://github.com/facebookresearch/fairseq/

[3] **GitHub:** https://github.com/C3Imaging/child_asr_conformer/

[4] **Conformer-transducer**: https://github.com/NVIDIA/NeMo/

TABLE III. WER FOR DIFFERENT NON-FINETUNED WHISPER, WAV2VEC2, AND CONFORMER-TRANSDUCER MODELS ON CHILD SPEECH (MYST, PFSTAR, AND CMU-KIDS) EVALUATION DATASETS

| Name | Models | MyST_test | PFS_test | CMU_test |
|---|---|---|---|---|
| Conformer-Transducer | Small | **21.34** | 12.68 | 16.05 |
| | Medium | 24.99 | 11.58 | 17.51 |
| | Large | 25.91 | 8.94 | 15.06 |
| | Xlarge | 24.42 | **8.22** | **14.83** |
| Whisper | Tiny | 40.09 | 159.57 | 30.63 |
| | Tiny.en | 33.02 | 47.11 | 27.32 |
| | Base | 32.14 | 100.07 | 25.03 |
| | Base.en | 29.15 | 45.70 | 20.75 |
| | Small | 26.22 | 111.75 | 18.52 |
| | Small.en | 26.72 | 39.00 | 16.82 |
| | Medium | 25.11 | 80.97 | **12.67** |
| | Medium.en | 28.06 | **35.25** | 14.00 |
| | Large | 25.24 | 84.52 | 13.70 |
| | Large-V2 | **25.00** | 73.68 | 12.69 |
| wav2vec2 | wav2vec2-base | 15.41 | 11.20 | 16.33 |
| | wav2vec2-large | **12.50** | **8.56** | **14.85** |

The Conformer-transducer Small model, which is significantly smaller than Whisper's Tiny model and five times smaller than Whisper's Base model outperforms both significantly on all three child audio evaluation sets. The Conformer-transducer Medium model, comparable in size to Whisper's Tiny model also outperforms Whisper but does not reach the same accuracy as wav2vec2-base, which is three times bigger, though the WERs are relatively close. The large Conformer-transducer model again outperforms the Whisper model of roughly the same parameter size range (Whisper Small) across all three child evaluation datasets and performs better on PFS_test and CMU_test than wav2vec2-base while its performance on MyST_test is significantly worse than wav2vec2-base. Conformer-transducer Xlarge model, which is twice the size of wav2vec2-large, only performs on par with it for PFS_test and CMU_test, while again showing a much poorer result on MyST_test. XLarge model, being slightly smaller in size than the Whisper Medium model, slightly outperforms the Whisper Medium model on MyST_test, significantly outperforms the Whisper Medium model on PFS_test, and does not outperform on CMU_test.

Overall, it can be observed that smaller Conformer-transducer models perform better than their small Whisper and wav2vec2 counterparts, while with an increase in parameter size, the Whisper and wav2vec2 models tend to outperform Conformer-transducer equivalents, suggesting that the Conformer-transducer loses its generalization capabilities with an increase in parameter size. The Conformer-transducer 'Large' and 'Xlarge' models demonstrated competitive performance in most cases. The Whisper models generally exhibited higher WERs compared to the Conformer-transducer models. However, the 'Medium' and 'Large' Whisper models showed impressive results on all three datasets. The wav2vec2 models, particularly the 'wav2vec2-large' model, achieved the lowest WERs among all the models evaluated.

*b)* **Comparative analysis between Conformer-transducer, Whisper, and wav2vec2 after finetuning:** Conformer-transducer finetuning experiments involved using the Large and Xlarge models, selected after analyzing the results of non-finetuned models on child evaluation datasets. Whisper finetuning included the Medium.en and Large-V2 models while wav2vec2 finetuning involved wav2vec2-base and wav2vec2-large models. These models were finetuned on MyST_55h, PFSTAR_10h, and a combination of both datasets.

TABLE IV. WER ON CHILD EVALUATION DATASETS FOR DIFFERENT WHISPER, WAV2VEC2, AND CONFORMER-TRANSDUCER MODELS FINETUNED ON MYST, PFSTAR, AND MYST+PFSTAR-COMBINED DATASETS

| Name | Models | MyST_test | PFS_test | CMU_test |
|---|---|---|---|---|
| **MyST (55 Hours) Finetuning:** | | | | |
| Conformer-Transducer | Large | 14.17 | 44.02 | 27.03 |
| | XLarge | 13.79 | 43.57 | 20.63 |
| Whisper | Medium.en | **11.81** | 17.83 | **15.07** |
| | Large-V2 | 12.28 | **10.88** | 15.67 |
| wav2vec2 | wav2vec2-base | 8.13 | 14.77 | 16.47 |
| | wav2vec2-large | **7.51** | **12.46** | **15.25** |
| **PFSTAR (10 Hours) Finetuning:** | | | | |
| Conformer-Transducer | Large | 90.00 | 8.58 | 82.00 |
| | XLarge | 86.79 | **6.31** | 75.26 |
| Whisper | Medium.en | 15.84 | 3.14 | 15.53 |
| | Large-V2 | **15.79** | **2.88** | 15.22 |
| wav2vec2 | wav2vec2-base | 31.86 | **3.48** | 27.49 |
| | wav2vec2-large | **27.17** | 3.50 | **21.35** |
| **MyST (55 Hours) + PFSTAR (10 Hours) Finetuning:** | | | | |
| Conformer-Transducer | Large | 13.86 | 4.44 | 25.00 |
| | XLarge | 13.61 | 4.3 | 21.21 |
| Whisper | Medium.en | 12.33 | 3.32 | 15.08 |
| | Large-V2 | 13.34 | 4.17 | 17.11 |
| wav2vec2 | wav2vec2-base | 7.94 | **2.91** | 15.97 |
| | wav2vec2-large | **7.42** | 2.99 | **14.18** |

A comparison between the Conformer-transducer, Whisper, and wav2vec2 WERs on the same evaluation sets can be seen in Table IV. First, a substantial increase in WER on the PFS_test and CMU_test is observed for the Conformer-transducer models finetuned on MyST_55h, while the WER on MyST_test is still higher than that for all Whisper and wav2vec2 models. Considering that CMU_test is the noisiest evaluation dataset, it is possible that, due to the higher WER of the Conformer-transducer on this set, the Conformer-transducer models deal worse with noisy datasets than the other model architectures. The results of the experiments with child speech finetuning show that wav2vec2 finetuning using MyST_55h resulted in lower WER compared to Whisper finetuning on MyST_test.

Finetuning the Conformer-transducer models on PFS_10h reduces the WER on PFS_test but again not to the same low levels as Whisper or wav2vec2 finetuning. Meanwhile, WERs on MyST_test and CMU_test is considerably higher for the Conformer-transducer models, again suggesting poor performance on noisier datasets. Finetuning the Conformer-transducer on a combination of MyST_55h and PFS_10h did not provide any improvements over the other models. However, when comparing to single dataset finetuning, the combined finetuning measurably improves the performance across all three evaluation datasets, suggesting that the model generalizes better when trained on more diverse and seen datasets.

Even though larger models tend to perform slightly better than their smaller counterparts, the performance gain from using larger models might not justify the additional computational cost and memory requirements, especially considering that the difference in WER between these models is relatively small. The performance of the models is heavily influenced by the finetuning dataset. Models finetuned on the MyST dataset tend to perform better on the MyST_test evaluation dataset, while those fine-tuned on the PFSTAR dataset achieve better results on the PFS_test evaluation dataset. This suggests that domain-specific finetuning is crucial for achieving better performance on domain-specific evaluation datasets.

Overall, the results in Table IV indicate that wav2vec2 may be the best ASR model for finetuning on child data, as the models are smaller and require drastically less data to train than Whisper models which show slightly poorer or comparable results at best. It consistently outperforms the other models across different finetuning datasets and evaluation datasets and achieves the lowest WER values for both MyST and PFSTAR datasets and their combination, indicating its effectiveness in capturing relevant speech features and generalizing to unseen data. While wav2vec2 shows promising results, it is important to note that the table might not cover all possible scenarios and datasets. Further evaluation and testing on different datasets would be required to validate the model's robustness and generalization capabilities.

## V. Conclusions

In this paper, the Conformer-transducer ASR model was compared against the Whisper and wav2vec2 models as approaches to improve the quality of child speech recognition. A fair comparison was conducted by ensuring that all models were evaluated within an identical parameter range and trained/evaluated using the same set of datasets. While the results show that finetuning the Conformer-transducer did not yield lower WER scores on child evaluation datasets compared to the Whisper or wav2vec2 finetuned models on the same datasets, there is still promise in using smaller-sized Conformer-transducer models for efficient low-resource deployment. The observed differences in finetuning performance may be attributed to the generalization capacity of the models, particularly for larger model sizes. It was evident that non-finetuned Conformer-transducer models had a more significant WER degradation compared to non-finetuned Whisper and wav2vec2 models as the model parameter size increased.

Furthermore, finetuned Conformer-transducer models perform worse on noisier evaluation datasets than Whisper and wav2vec2 models. Using a combination of datasets for finetuning improved WER scores across all datasets for the Conformer-transducer, suggesting that a more diverse finetuning dataset is needed for the model to generalize well to unseen data. On the other hand, when comparing non-finetuned models at smaller sizes, the Conformer-transducer model outperformed both the Whisper and wav2vec2 models within a similar parameter range across all child evaluation datasets. This indicates that Conformer-transducer models perform optimally at smaller sizes but may face challenges in maintaining generalization capabilities as their size increases. Overall, wav2vec2 showed the most promising results and can be considered to be the best ASR model for finetuning child data among the other models.

In future work, it is proposed to finetune the smaller Conformer-transducer models, namely Small and Medium, on child datasets. Additionally, more rigorous hyperparameter sweeping could provide lower WER scores as well as testing different decoding strategies such as beam-search with Time Synchronous Decoding (TSD) [32] or Alignment-Length Synchronous Decoding (ALSD) [33]. Finally, using different vocabulary sizes for the tokenizer may be investigated.

## Acknowledgment

The authors would like to acknowledge experts from Xperi Ireland: Gabriel Costache, Zoran Fejzo, and George Sterpu for providing their expertise and feedback while working on this research.

## Appendix

As of date, the only approach to finetuning Conformer-transducer models that are documented is simply training all layers of the encoder, decoder, and joint networks. However, we considered the possibility of finding a more optimal approach to finetuning which would lead to lower WER scores on the evaluation datasets. To determine the best combination of hyperparameters and what layers of the Conformer-transducer model to finetune for the main experiments detailed in Section IV.C, the large model was preliminarily fine-tuned on MyST_55h and the setup with the lowest WER on MyST_test was chosen as the finetuning approach to use.

The first approach involved finetuning all layers of all networks with the baseline hyperparameters recommended by the training scripts, which use the Adam optimizer with a learning rate of 5.0 and the Noam learning rate scheduler with 10,000 warmup steps. The lowest WER achieved on the MyST_test for this approach was 18.58%. The next approach modified the learning rate to 2.0, which led to a decreased WER of 16.3%. Further decreasing the learning rate to 1.0 achieved a WER of 14.55%. The next investigated approach involved finetuning just the feed-forward layers of the encoder network while freezing all other encoder layers, with a 1.0 learning rate and 10,000 Noam warmup steps, achieving a 14.21% WER. Using the Noam Hold learning rate scheduler with a warmup of 10,000 steps and a hold of 20,000 steps did not lead to improvements in WER on MyST_test. Finetuning only the final half of the feed-forward layers of the encoder instead of all the feed-forward layers also did not yield improvements. Finally, the best WER of 14.17% was achieved by finetuning all the feed-forward layers of the encoder with a learning rate of 3.0 and a Noam warmup of 40,000 steps. Note that all layers of the decoder and joint networks were fine-tuned in all of the preliminary experiments.